\documentclass[runningheads]{llncs}

\usepackage{eccv}

\usepackage{eccvabbrv}

\usepackage{graphicx}
\usepackage{booktabs}

\usepackage{bbding}

\usepackage[accsupp]{axessibility}  

\usepackage{hyperref}

\usepackage{orcidlink}

\usepackage{mwe}

\begin{document}

\title{What Emerges and What Breaks in Self-Play Driving}


\author{Laur Sisask\orcidlink{0009-0005-0285-6325} \and
Ardi Tampuu\orcidlink{0000-0002-2120-1712} \and
Tambet Matiisen\orcidlink{0000-0002-6815-1580}\Envelope}

\authorrunning{L.~Sisask et al.}

\institute{University of Tartu, Tartu, Estonia \\
\email{\{laur.sisask,ardi.tampuu,tambet.matiisen\}@ut.ee}}

\maketitle

\begin{abstract}
  Training autonomous driving policies through pure self-play has recently shown promising results. Following Gigaflow and PufferDrive, we train driving policies in a similar self-play fashion, but extend the models from MLPs to Transformers and train on the high-definition map of a real city, where we ultimately aim to deploy them. On the CARLA and Waymax benchmarks, our policies fall short of Gigaflow, and we trace the gap to specific failure modes, including reward hacking at traffic lights and a missing incentive to stop at stop signs. We further analyze which traffic rules emerge from self-play and how closely they match human driving, and we confirm that reward conditioning yields the intended diversity of driving behaviors. A demonstration of a trained policy is available at \url{https://laursisask-ut.github.io/eccvdemo}.
\end{abstract}

\section{Introduction}
\label{sec:intro}

Learning to drive through reinforcement learning requires far more experience than can safely be collected on real roads. Driving simulators are a natural alternative to real-world data collection. Photorealistic simulators such as CARLA model the driving environment in detail but have relatively low throughput, making them less suitable for large-scale reinforcement learning. Over the last few years, a number of driving simulators have emerged that meet the need for collecting large amounts of data by simulating traffic at a high level of abstraction, representing the world as objects such as vehicles and lanes instead of raw sensor data. Simulators such as Gigaflow~\cite{gigaflow}, PufferDrive~\cite{pufferdrive2025github}, and Waymax~\cite{waymax} make it possible to train driving policies at large scale, with or without human driving examples, and prior work has shown that the resulting policies can become reasonably good drivers~\cite{gigaflow,reliablesim}. Most notably, Gigaflow reported more than 3 million simulated km between collision or off-road incidents in a long-form self-play evaluation~\cite{gigaflow}. Motivated by these results, we set out to build a similar system with the end goal of deploying it in the real world, in Tartu, Estonia.

In this article, we work towards reproducing the results of Gigaflow using the PufferDrive software library. Rather than attempting to reproduce Gigaflow's benchmark scores, we investigate whether its broad self-play approach can support our longer-term goal of deployment in Tartu. We make the following major changes to the training process. First, we train on the high-definition map of Tartu instead of synthetic maps, which brings training closer to our intended deployment domain. Second, we explore Transformer architectures in addition to the MLPs used in prior work. Although our models do not reach the performance of Gigaflow, we analyze in detail where and why they fail, study which traffic rules the policies learn and where their behavior diverges from human driving, and verify that the reward conditioning produces the intended diversity of behaviors.

\section{Simulation Environment}

We train our models with PufferDrive~\cite{pufferdrive2025github}, a high-throughput autonomous driving simulator and training loop that combines self-play with the proximal policy optimization algorithm~\cite{ppo}. We modify the simulator in several ways, described in the sections that follow and for the most part inspired by Gigaflow~\cite{gigaflow}.

The environment runs at a timestep of $0.1$ seconds, with episodes lasting 20, 30, or 40 seconds. Although training is episodic, we treat the task as continuous by bootstrapping the return from the final state with the learned value function.

\subsection{The Map}

We train our models in a world built from a high-definition map of Tartu, Estonia, a mid-sized European city. Since our ultimate goal is to deploy the models on a real autonomous vehicle in Tartu, training on the same map keeps the training environment close to the deployment environment. This contrasts with Gigaflow~\cite{gigaflow}, which used synthetic CARLA maps~\cite{carla}, and with Cornelisse~\etal~\cite{reliablesim}, which used real map data from the United States.

Our map consists of 436 km of drivable lanes and presents diverse traffic scenarios, from small residential roads with right-hand-rule intersections to roundabouts and regulated intersections with up to 16 incoming lanes in total. The drivable area is captured by a set of lines indicating the edges of the road, and lanes are represented by their centerlines. Regulatory features such as traffic lights, yield signs, stop signs, and crosswalks are all represented by special lines on the map, each marking the position where a vehicle must stop when required. One limitation is that our map represents yield- and stop-sign lines as a single entity, which we call a stop line. Because yield signs are much more common in Tartu, we treat all stop lines as yield lines. This means that we do not desire that an autonomous vehicle stops at these lines unless it needs to yield to another vehicle.

\begin{figure}[tb]
  \centering
  \begin{subfigure}{0.32\linewidth}
    \includegraphics[width=\textwidth]{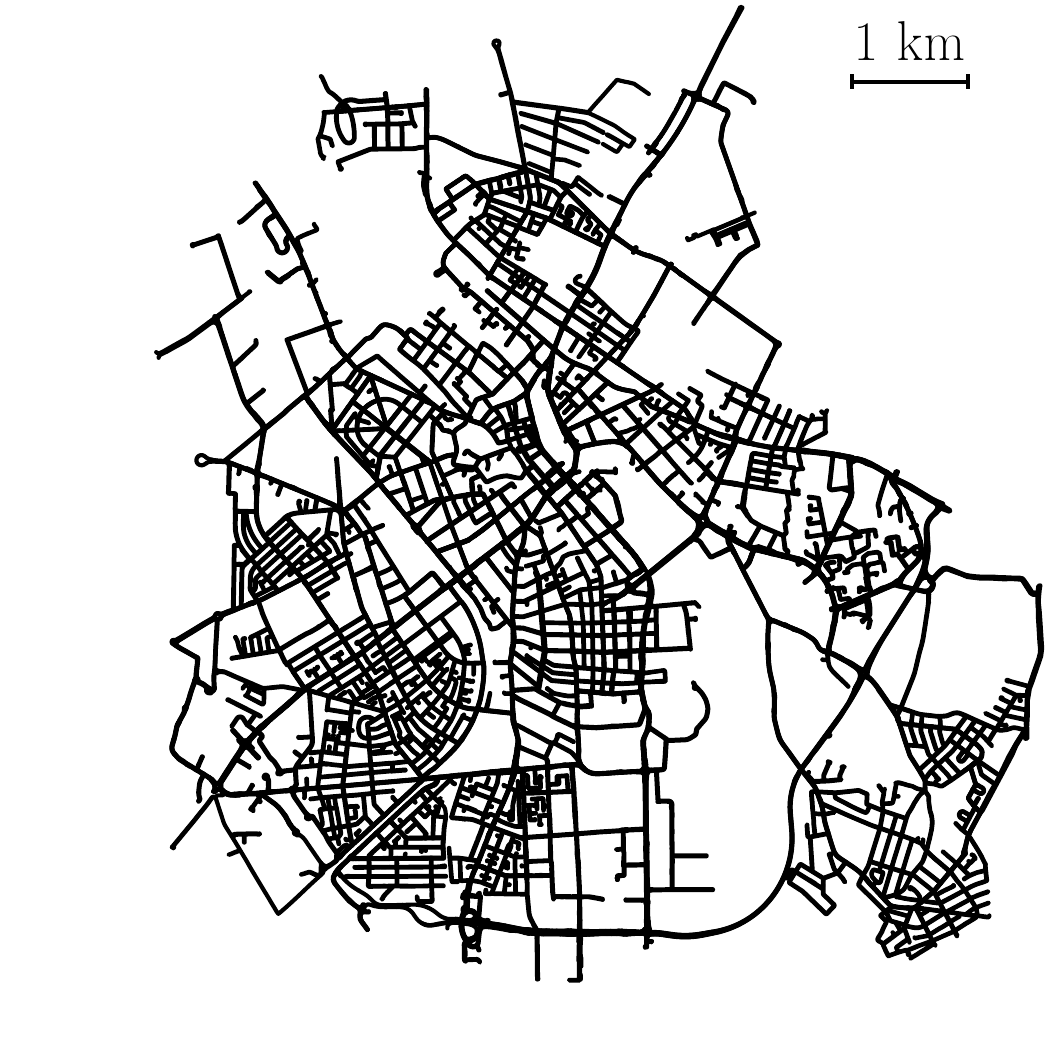}
    \caption{High-level structure of the map}
  \end{subfigure}
  \hfill
  \begin{subfigure}{0.32\linewidth}
    \includegraphics[width=\textwidth]{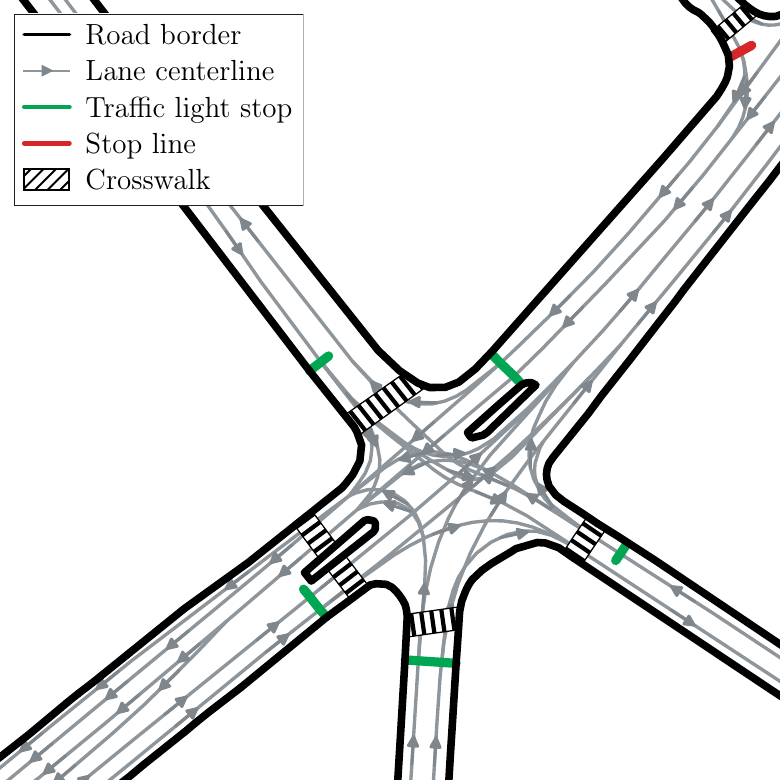}
    \caption{An example of a regulated intersection}
  \end{subfigure}
  \hfill
  \begin{subfigure}{0.32\linewidth}
    \includegraphics[width=\textwidth]{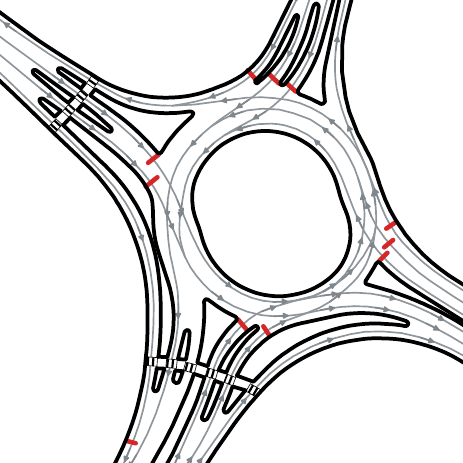}
    \caption{An example of a roundabout}
  \end{subfigure}
  \caption{HD map of Tartu used in the training}
  \label{fig:map}
\end{figure}

The lanes form a graph in which one lane is connected to another if a vehicle is legally allowed to drive from the first to the second. This graph is used to plan a global route between any two points on the map. The map is shown in \cref{fig:map}.

\subsection{Initialization}

At the start of each training episode, we sample a small region of our map: a $200 \times 200$ m area in which we randomly spawn 32 agents, requiring at least 15 meters between any two vehicles. Each vehicle receives randomly sampled dimensions, ranging from a bicycle to a bus (more details are in supplementary material). 90\% of the agents spawn aligned with the lane direction, driving at a low randomly sampled initial speed, while the remaining 10\% start as parked vehicles on the side of the road. Finally, we use the lane graph to generate up to 8 waypoints per agent, spaced on average 5 seconds apart when driving at the applicable speed limit.

\begin{figure}[tb]
  \centering
  \begin{subfigure}{0.48\linewidth}
    \includegraphics[width=\textwidth]{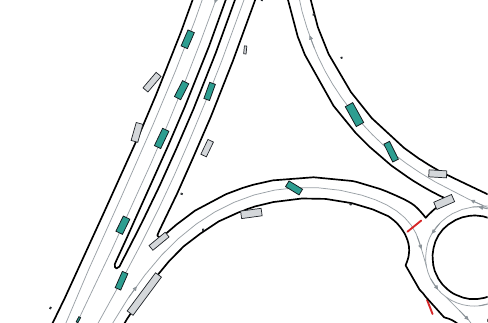}
  \end{subfigure}
  \hfill
  \begin{subfigure}{0.48\linewidth}
    \includegraphics[width=\textwidth]{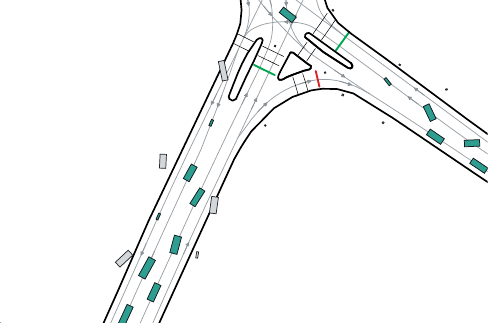}
  \end{subfigure}
  \caption{Examples of initial vehicle positions. Green vehicles are moving vehicles controlled by the policy, and grey vehicles are stationary. The small square boxes are pedestrians following fixed trajectories.}
  \label{fig:init}
\end{figure}

After the spawn locations have been determined, we additionally place up to 48 stationary vehicles near the side of the road, as well as pedestrians that cross the road by following precomputed trajectories, both at crosswalks and elsewhere. For traffic lights, we use traffic-light randomization inspired by Gigaflow but with a different distribution: all traffic lights are disabled in 50\% of the episodes, and otherwise each light is initialized with a random state (off, red, yellow, or green). During the episode, the lights change state at random times, with each state lasting between 0.5 and 10 seconds. \Cref{fig:init} shows two examples of initial episode states.
\subsection{Reward Function and Conditioning}
\label{sec:rewards}

Following Gigaflow~\cite{gigaflow}, our agents receive the following reward terms, most of which have independently sampled coefficients supplied to the model as inputs:

\begin{itemize}
    \item a reward for reaching a waypoint, given when the agent's distance from the waypoint is less than $D$ m, always $R = 0.25$, $D \sim U(1, 3)$
    \item a penalty for colliding with another vehicle, given as $R = -C_{\text{collision}} (1 + 0.1v)$ where $v$ is speed of the vehicle in m/s, $C_{\text{collision}} \sim U(0, 1)$.
    \item a penalty for colliding with the road border, given as $R = -C_\text{offroad}(1+0.1v)$ where  $v$ is speed of the vehicle in m/s, $C_\text{offroad} \sim U(0, 1)$
    \item a penalty for crossing a traffic light line while it is red, given as $R = -C_\text{red}$, $C_{\text{red}} \sim U(0, 1)$
    \item a penalty at every timestep where longitudinal or lateral acceleration exceeds $3 \: \textrm{m}/\textrm{s}^2$, given as $R = -C_\textrm{comfort}$, $C_\textrm{comfort} \sim U(0, 0.01)$
    \item a reward for driving close to center of road, given at every timestep when angle between vehicle and road centerline is less than $60$ degrees, $R = C_\textrm{center} (-\left| B - s \right| + \frac{0.05}{\textrm{exp}(\left| B - s \right| - 0.5)})$ where $B$ is bias and $s$ distance in meters from road center, $C_{\textrm{center}} \sim U(2.5\cdot10^{-5},\,2.5\cdot10^{-4})$ with probability $50\%$, and $C_{\textrm{center}} \sim U(2.5\cdot10^{-4},2.5\cdot10^{-3})$ with probability $50\%$, $B \sim U(-0.3, 0.3)$.
    \item a reward at every timestep based on the angle between the agent and the road, given as $R = C_\text{align} (\min(\cos \alpha, 0) + 0.0025\cdot (1- \left|\alpha\right| / \frac{\pi}{2}))$ where $\alpha$ is angle between the agent and centerline in radians, $C_{\textrm{align}} \sim U(2.5\cdot10^{-5},\,2.5\cdot10^{-3})$ with probability $50\%$, and $C_{\textrm{align}} \sim U(2.5\cdot10^{-3},2.5\cdot10^{-2})$ with probability $50\%$
    \item a reward at every timestep where vehicle is moving at least $2.5 \:\textrm{m}/\textrm{s}$, given as $R = 2.5\cdot10^{-4} \cdot \max(\cos\alpha, 0)$ where $\alpha$ is the angle between the vehicle and road centerline. 
\end{itemize}

A vehicle collision terminates an agent's trajectory when $C_\text{collision}\geq0.5$, and a road-border collision does so when $C_\text{offroad}\geq0.5$. This mirrors the real world, where a drive ends after a collision and no further rewards can be collected. 

We do not penalize agents for running over stop lines. This should work identically to Gigaflow, but we note that the Gigaflow article makes some conflicting claims regarding reward at stop lines, so we are not fully sure about this. Regardless, our main reason to not penalize at stop lines is due to our treatment of them as yield signs, not stop signs --- we do not want our vehicle to stop at a stop line unless it is to yield to another vehicle.

The purpose of this reward randomization is to populate the environment with a diverse set of agents. For example, some drivers avoid collisions at all costs, while others largely ignore them and focus on keeping to the center of their lane.

\subsection{Action Space}
We use the discrete action space of PufferDrive with two action dimensions --- one for the steering angle and one for acceleration. For acceleration, we use the PufferDrive defaults with values $\pm\{0, 1.33, 2.67, 4\} \: \textrm{m/s}^2$. For steering, we diverge from the defaults: early experiments showed that the agents struggled to keep the car aligned with the road because the default action space contained only sharp turning angles. We therefore replace some of the sharpest angles with smaller ones, giving the policy more granular control over the vehicle. The resulting steering space is $\pm\{0, 2, 5, 10, 19, 29, 38\} \: \textrm{degrees}$.

\subsection{Observation Space}

We start with the default PufferDrive observation space, which includes the positions and speeds of the closest 63 objects and the positions of the closest 200 road elements, both within 100 meters. Since our perception stack in the real world does not distinguish between classes of objects like pedestrians and vehicles, we do not include the class in the observation space. If needed, the model can infer the class of an object based on its properties like dimensions or speed.

Each road element is a straight line segment on the map, observed through its middle point and direction relative to the ego vehicle. For traffic light lines, the ego additionally observes its current state --- off, red, yellow, or green. In addition, the ego vehicle observes its own dimensions, speed, current and next waypoint, the number of remaining waypoints, its distance from the lane center, the angle between the lane and the vehicle, and the reward conditioning coefficients. The observation space is illustrated in \cref{fig:observationspace}.

\begin{figure}[tb]
  \centering
\includegraphics[width=0.7\textwidth]{observation_space_beautiful.pdf}
  \caption{Observation space of the ego vehicle.\\\textbf{Road lines.} Road edges and lane centerlines are represented by segments, each of which is a straight line. The ego has access to the position of the segment's center point in egocentric coordinates ($dx$, $dy$), the angle between the line and the ego's direction ($\theta$), and the segment's length.\\\textbf{Other vehicles.} The ego vehicle has access to the position of each vehicle in egocentric coordinates ($dx$, $dy$), the angle between the ego's direction and the vehicle's direction ($\theta$), as well as the vehicle's length, width, and velocity ($\vec{v}$).\\\textbf{Ego state.} The ego can access its own width, length, velocity ($\vec{v}$), the angle between the lane centerline and the vehicle ($\alpha$), its lateral distance from the centerline ($s$), and positions of the next two waypoints in egocentric coordinates ($dx$, $dy$).}
  \label{fig:observationspace}
\end{figure}

\subsection{Physics}

In our early experiments, agents drove through sharp curves at speeds that would be impossible in the real world or in more realistic driving simulators. We therefore extend the PufferDrive simulator to limit the friction available to the vehicles. For each vehicle, we sample a coefficient of friction $\mu \sim U(0.7, 1.3)$. Whenever the agent's desired net acceleration exceeds $\mu g$, where $g = 9.81 \: \textrm{m/s}^2$, we constrain the lateral and longitudinal acceleration applied to the vehicle. As a result, a vehicle entering a turn too fast understeers and follows a trajectory with a larger radius than the policy requested. Unlike the other observed variables, the friction coefficient is hidden from the ego vehicle.

In addition, we introduce two further hidden variables: a steering gain and an acceleration gain, each sampled from $U(0.9, 1.1)$. Every action selected by the policy is multiplied by the corresponding gain. Gigaflow used a similar form of conditioning, giving different vehicles different steering and acceleration capabilities, but in our case the gains are hidden from the model, whereas they included them in the agent observation.

These two variables proved crucial for running our models in environments such as CARLA, where acceleration and steering are not controlled directly and follow more complex dynamics. Without this variation, our models failed to make any meaningful progress in CARLA due to excessive or insufficient steering.

\section{Models}

\begin{figure}[tb]
  \centering
  \includegraphics[width=0.77\textwidth]{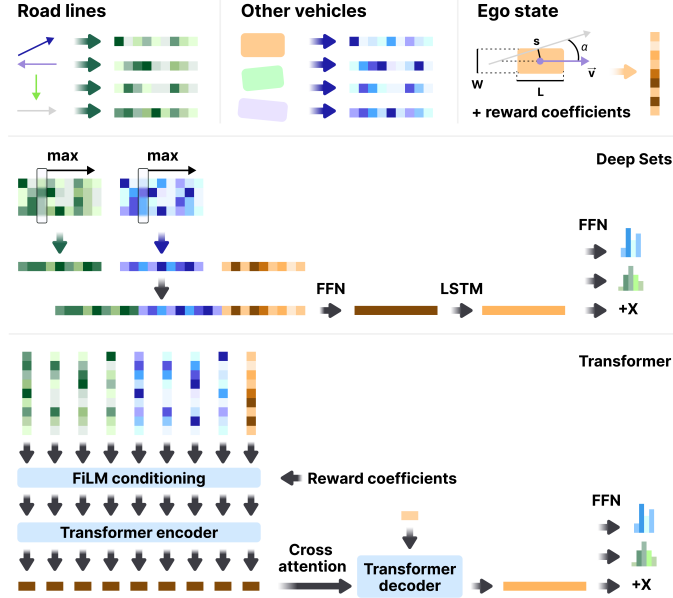}
  \caption{Model architectures}
  \label{fig:models}
\end{figure}

We train two models: a model from PufferDrive based on the Deep Sets architecture~\cite{deepsets}, and a Transformer~\cite{transformer} inspired by Wayformer~\cite{wayformer}. Both models use separate networks to encode observation data from the different modalities into sets of vectors of the same size. Both architectures are illustrated in \cref{fig:models}.

The Deep Sets model collapses each set into a fixed-size, order-invariant vector by taking an element-wise maximum over the set. This is done for each modality separately, and the resulting vectors are fed into an MLP followed by an LSTM, with linear heads producing the actor and critic outputs. The architecture shares Gigaflow's modality-specific encoding and max pooling, but ours adds an LSTM. Additionally, Gigaflow uses separate feed-forward actor and critic networks.

The Wayformer-based model likewise converts each data point in the observation into a fixed-size vector. These vectors are modulated by the reward conditioning coefficients through a FiLM layer~\cite{film} and then become input tokens to a Transformer encoder, where self-attention is allowed between all tokens, including those from different modalities. A Transformer decoder with a single learned query embedding and cross-attention over the encoder outputs produces the actor and critic outputs. Both Transformer encoder and decoder have layer normalization at the start of every sub-layer and use ReLU as activation function. Unlike the original Wayformer, our Transformer only accesses the current state, a simplification that lowers the computational cost of running the model.

\section{Experiments}

The Deep Sets model uses the default configuration from PufferDrive: all inputs are projected to vectors of dimensionality 64 and the LSTM has a hidden size of 256, giving 596K parameters in total. The Transformer uses embeddings of dimensionality 128, an encoder with two Transformer blocks, a decoder with one block, and four attention heads, giving 638K parameters. Both models are thus around 10$\times$ smaller than the Gigaflow model, which has around 6 million parameters. We experimented with larger models but settled on these small ones: in preliminary experiments, model size made little difference in capability, and smaller models train with higher throughput.

We use the default training loop from PufferDrive with the PPO-clip algorithm~\cite{ppo}. Each model trains for 8--10 days on a single NVIDIA B200 GPU and 32 CPU cores, with 16,384 parallel environments for the Deep Sets model and 4,096 for the Transformer. Other hyperparameters and learning curves are given in supplementary material. Over the course of training, the Deep Sets model observes 630 years' worth of driving data and the Transformer 57 years' worth. The comparison is therefore matched by hardware and training time rather than by environment transitions.

\section{Results}

We evaluate our models on the existing CARLA~\cite{carla} and Waymax~\cite{waymax} benchmarks, as well as on our own custom scenarios. While the vehicles train with diverse reward conditioning, evaluation uses fixed coefficients. For all reward terms, we set all reward coefficients to the maximal values used during training. The bias coefficient for lane center we set to 0 and goal radius to 3 m.

\subsection{CARLA}

Following Gigaflow, we use privileged traffic-participant states and detailed map geometry. Since CARLA route outcomes vary with the random seed, we run each route three times. \Cref{tab:carla} summarizes the results on CARLA. Overall, our models score worse than Gigaflow. We attribute this to several factors, which we analyze below.

\begin{table}[tb]
  \caption{Results on different CARLA benchmarks. DS stands for Driving Score, RC for Route Completion, IP for Infraction Penalty, Ped for collisions with
  pedestrians, Veh for collisions with vehicles, Lay for collisions with layout, Red for red light violations, Stop for stop sign violations, Off for driving off-
  road, Dev for deviating from the route, TO for timeout, and Block for the vehicle getting blocked.}
  \label{tab:carla}
  \centering
  \setlength{\tabcolsep}{2pt}
  \begin{tabular}{lcccccccccccc}
  \toprule
  Model & DS & RC & IP & Ped & Veh & Lay & Red & Stop & Off & Dev & TO & Block \\ \midrule
  \multicolumn{13}{l}{\textit{CARLA Leaderboard 1.0 Testing Routes} \cite{carla}} \\
  Deep Sets & $30{\pm}1$ & $56{\pm}5$ & $0.60{\pm}0.05$ & $0.03$ & $0.16$ & $0.26$ & $1.18$ & $0.55$ & $3.28$ & $0.20$ & $0.03$ & $0.60$ \\
  Transformer & $29{\pm}2$ & $84{\pm}4$ & $0.41{\pm}0.05$ & $0.00$ & $0.02$ & $0.11$ & $2.20$ & $0.73$ & $0.72$ & $0.11$ & $0.01$ & $0.14$  \\
  Gigaflow & $93{\pm}1$ & $97{\pm}2$ & $0.95{\pm}0.01$ & $0.00$ & $0.07$ & $0.00$ & $0.01$ & $0.00$ & $0.64$ & $0.01$ & $0.07$ & $0.00$ \\ \midrule
  \multicolumn{13}{l}{\textit{LAV benchmark} \cite{lav}} \\
  Deep Sets & $27{\pm}3$ & $52{\pm}9$ & $0.62{\pm}0.04$ & $0.00$ & $0.10$ & $0.00$ & $1.12$ & $1.91$ & $2.61$ & $0.26$ & $0.00$ & $0.73$ \\
  Transformer & $17{\pm}2$ & $80{\pm}1$ & $0.34{\pm}0.00$ & $0.00$ & $0.04$ & $0.00$ & $3.66$ & $2.13$ & $0.04$ & $0.09$ & $0.00$ & $0.30$  \\
  Gigaflow & $99{\pm}1$ & $99{\pm}1$ & $1.00{\pm}0.00$ & $0.00$ & $0.00$ & $0.00$ & $0.00$ & $0.00$ & $0.00$ & $0.00$ & $0.02$ & $0.00$  \\ \midrule
  \multicolumn{13}{l}{\textit{Longest6 benchmark} \cite{transfuser}} \\
  Deep Sets & $18{\pm}1$ & $36{\pm}1$ & $0.62{\pm}0.02$ & $0.06$ & $0.38$ & $0.17$ & $0.18$ & $0.46$ & $3.99$ & $0.42$ & $0.02$ & $1.36$ \\
  Transformer & $23{\pm}0.3$ & $67{\pm}1$ & $0.44{\pm}0.00$ & $0.01$ & $0.10$ & $0.03$ & $2.82$ & $0.64$ & $0.86$ & $0.18$ & $0.03$ & $0.40$  \\
  Gigaflow & $92{\pm}2$ & $99{\pm}1$ & $0.93{\pm}0.01$ & $0.02$ & $0.08$ & $0.00$ & $0.04$ & $0.03$ & $0.24$ & $0.03$ & $0.05$ & $0.00$  \\ \bottomrule
  \end{tabular}
  \end{table}

\begin{figure}[tb]
  \centering
  \begin{subfigure}{0.32\linewidth}
    \includegraphics[width=\textwidth]{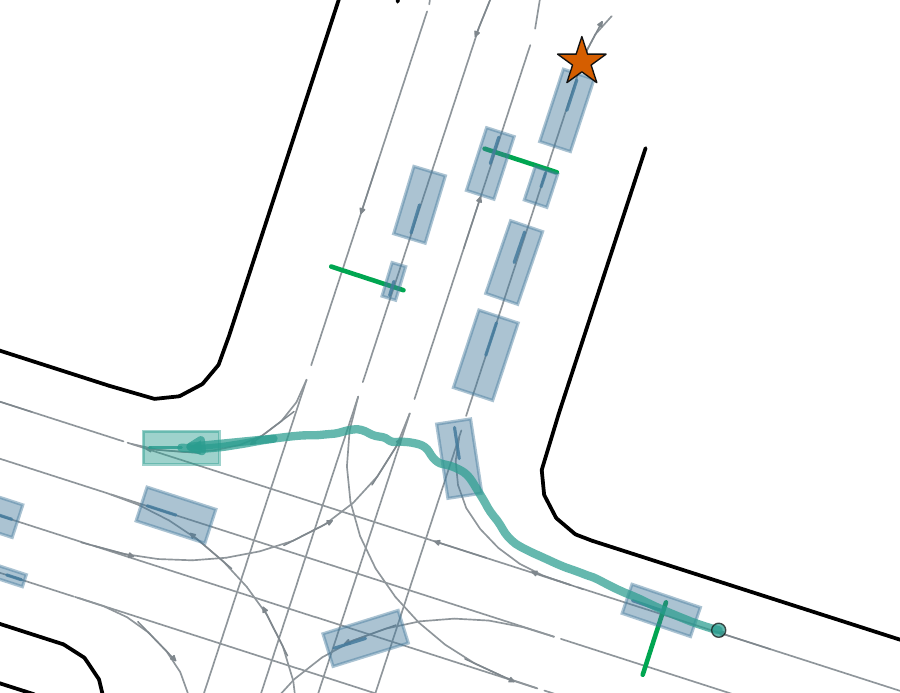}
    \caption{Deviation from the route}
    \label{fig:carla-failures-deviation}
  \end{subfigure}
  \hfill
  \begin{subfigure}{0.32\linewidth}
    \includegraphics[width=\textwidth]{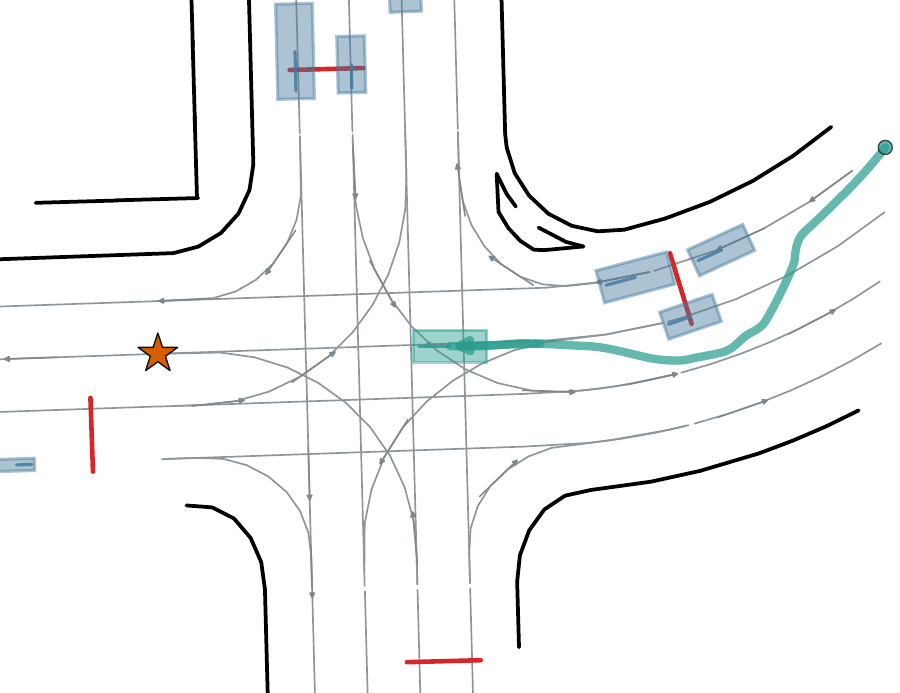}
    \caption{Reward hacking at red light}
    \label{fig:carla-failures-redlight}
  \end{subfigure}
  \hfill
  \begin{subfigure}{0.32\linewidth}
    \includegraphics[width=\textwidth]{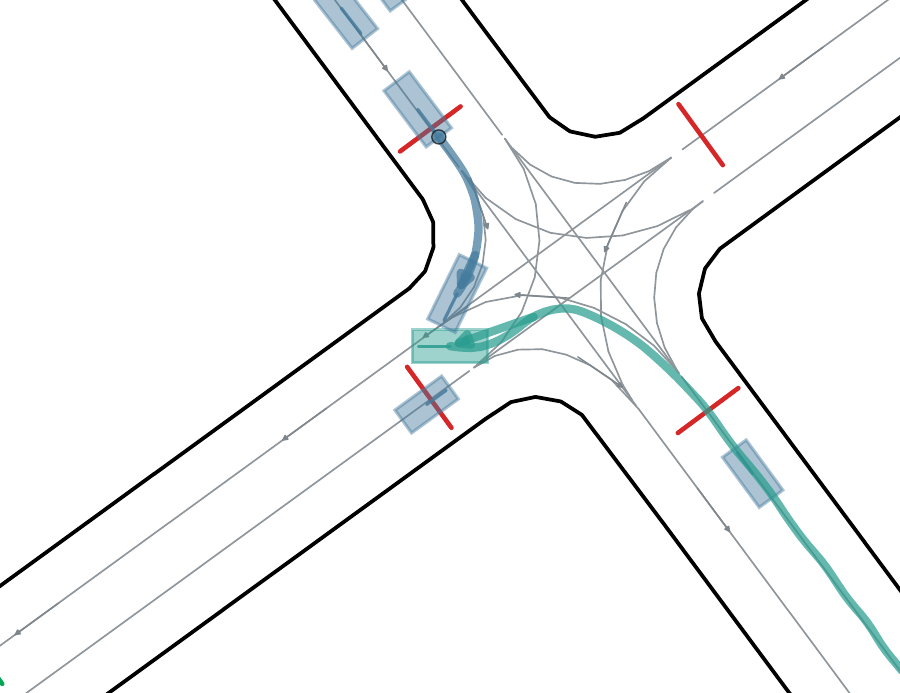}
    \caption{Collision with a vehicle}
    \label{fig:carla-failures-collision}
  \end{subfigure}
  \caption{Examples of failures on CARLA routes. Yellow star denotes the position of the waypoint. Green shade shows the trajectory taken by the policy. The NPC vehicles are shown in blue.}
  \label{fig:carla-failures}
\end{figure}

\subsubsection{Completion Rate.} There are two reasons why our policies do not reach a 100\% completion rate --- deviating from the route (occurs in 45\% of the scenarios) and becoming blocked (in 17\% of the scenarios). Deviations from the route occur when the policy drives in the wrong direction at an intersection. After reviewing some scenarios, one case where this frequently happens seems to be when the path to the waypoint is blocked by stopped vehicles, as shown in \cref{fig:carla-failures-deviation} where the vehicle started turning to the right towards a waypoint, but diverted back to the main road.

The other reason, becoming blocked, occurs in two situations. The first is when our policy attempts to pass a vehicle in front of it while another vehicle is driving towards it: both vehicles brake before colliding, and neither our policy nor the NPC takes action to resolve the situation. The second case is tight turns where the policy stops in the middle of the turn even if there is space to complete it. We hypothesize that this happens due to such turns being out-of-distribution for our models, as unlike Gigaflow, they were not trained on CARLA maps.

\subsubsection{Stop Signs.} Stop-sign violations are common. We hypothesize this happens because the policy lacks a reward signal for stopping at stop signs (for reasons explained in~\cref{sec:rewards}). We watched recordings of 117 stop sign violations and found that our policy does stop at stop signs, but only when another vehicle is approaching the intersection. More specifically, we found that in 62\% of violations, there were either no other vehicles at an intersection or the vehicle did not need to interact with them (\eg turning right when vehicle approaches from right). In 24\% of the cases, the policy did stop to yield to another vehicle, but did so either before or after the stop sign. In the remaining 14\% of cases, the policy did not stop and created dangerous situations. 

\subsubsection{Red Lights.} After manually inspecting 110 traffic light violations of our Transformer model, we estimate that around 84\% happened during traffic light transitions from green to yellow to red after the vehicle had already crossed the red light stop line. We hypothesize that these violations occur because of a discrepancy between our training environment and CARLA --- we penalize the agent if its bounding box intersects with a red stop line, while CARLA penalizes the agent if it has not fully driven into the intersection when light turns red, even if the agent has crossed the stop line. 

In the remaining 16\% of violations, our model engaged in reward hacking. In our environment, traffic lights are modeled as colored stop lines, and agents are penalized for crossing them while they are red. At intersections, these stop lines only span the lane entering the intersection, which allows the vehicle to reward hack by using the oncoming lane to bypass the red stop line without incurring a penalty. An example of this is shown in \cref{fig:carla-failures-redlight}. Interestingly, Gigaflow incurred very few red light violations, which is interesting because, to the best of our knowledge, they modeled the traffic light behavior and reward very similar to us.

\subsubsection{Collisions.} When it comes to collisions with other vehicles and pedestrians, our Transformer model achieves performance broadly similar to Gigaflow across the three benchmarks. In total, the Transformer-based policy participates in 15 collisions with other vehicles and one with a pedestrian. Two of these collisions happen due to the policy not yielding at a four-way stop, one of which is shown in \cref{fig:carla-failures-collision}. Three collisions happen while vehicle drives in oncoming lane while passing another vehicle, five are side-swipes, in three cases the vehicle is rear-ended by an NPC, one is a head-on collision caused by NPC driving into oncoming traffic, and in one scenario the route starts with vehicle already in collision. The sole collision with pedestrian happened when a pedestrian spawned on the road, close to the vehicle, leaving very little time to react. As for collisions with the layout, we found it difficult to model some objects such as light poles, as the CARLA simulator only exposes bounding boxes, which for light poles extend over the lane. We did not include these objects in the observations, which led to occasional collisions with them.

\subsection{Waymax}

Similar to Gigaflow, we run our evaluation on scenarios from the validation split of the Waymo Open Motion Dataset (WOMD) version 1.2.0~\cite{womd}. The ego vehicle is controlled by our policy, and the other vehicles are controlled by the IDM algorithm~\cite{idm}. \Cref{tab:waymax} summarizes the results on Waymax.

On Waymax, our models perform worse than Gigaflow on the collision and off-road rate metrics. We looked at 56 failures and all these failures can be attributed to one of the following causes:

\begin{itemize}
    \item Policy drives too close to the road edge and hits it (18\%)
    \item While driving close to the road edge, the ego vehicle turns away from the edge too sharply and its rear end hits the road edge (14\%)
    \item The ego vehicle does not leave enough room to the side when passing another vehicle (13\%)
    \item The vehicle in front or coming from the side brakes, and the ego policy hits it (14\%)
    \item A vehicle spawns in front or inside of the ego vehicle, and the policy collides with it (20\%)
    \item The ego vehicle invades the lane of another vehicle, and that vehicle hits the ego from the side or from behind (11\%)
    \item Other failures (10\%)
\end{itemize}

\begin{table}[tb]
\caption{Results on the Waymax benchmark on the WOMD validation set. One vehicle is controlled by our model and the others by IDM~\cite{idm}. Score is Gigaflow's custom aggregate, not an official Waymax metric.}
\label{tab:waymax}
\centering
\begin{tabular}{cccccccc}
\toprule
Model & Off-road & Collision & Kinematic & Log & Route & Total & Score \\
& rate & rate & infeasibility & ADE & progress & driven ratio & \\ \midrule
Expert demo & 0.32\% & 0.61\% & 4.33\% & 0.00 m & 100\% & 100\% & $\leq 99.07\%$ \\ \midrule
Deep Sets & 2.36\% & 4.32\% & 0.10\% & 11.38 m& 207\% & 137\% & 93.61\% \\
Transformer & 1.70\% & 2.20\% & 0.10\% & 8.34 m& 162\% & 120\% & 96.35\% \\
Gigaflow & 0.43\% & 0.43\% & 0.14\% & 5.87 m & 146\% & 107\% & 99.16\% \\ \bottomrule
\end{tabular}
\end{table}

\subsection{Adherence to Traffic Rules}

\begin{figure}[tb]
  \centering
  \begin{subfigure}{0.4\linewidth}
    \includegraphics[width=\textwidth]{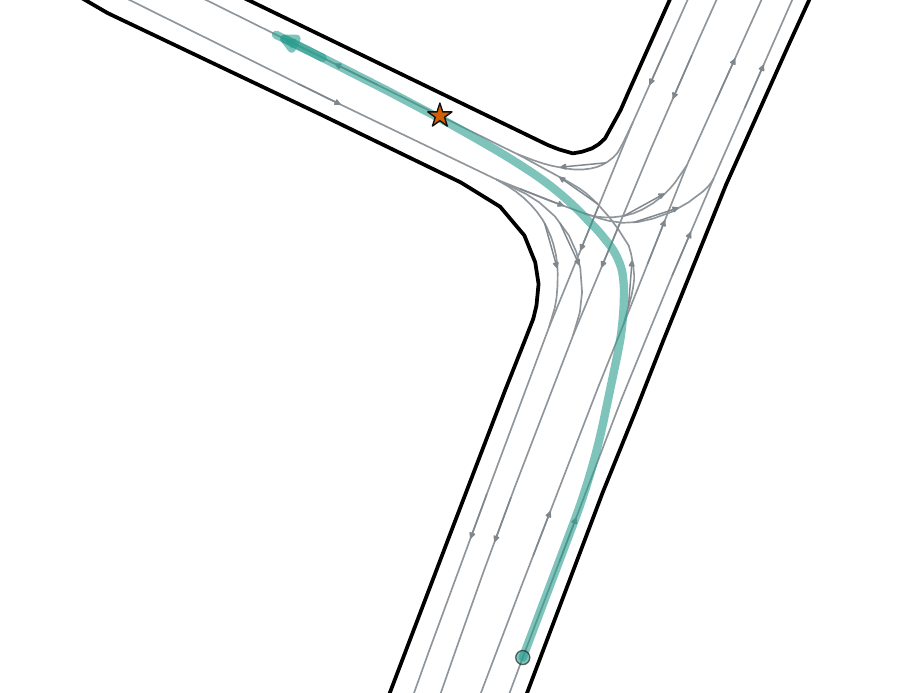}
    \caption{Left turn}
    \label{fig:rules-leftturn}
  \end{subfigure}
  \hfill
    \begin{subfigure}{0.4\linewidth}
    \includegraphics[width=\textwidth]{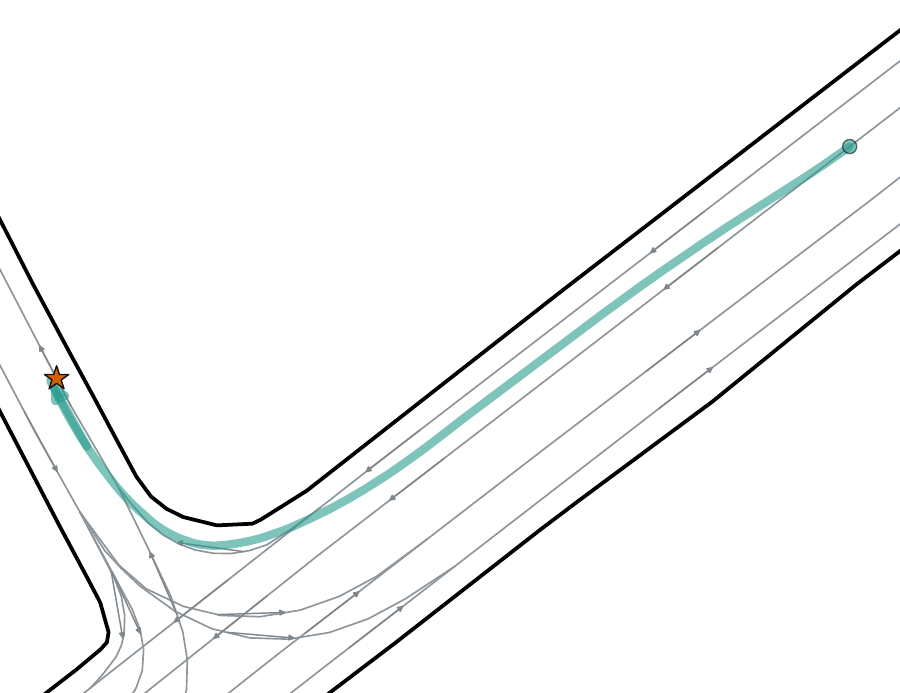}
    \caption{Right turn}
    \label{fig:rules-rightturn}
  \end{subfigure}
  \caption{Examples of synthetic scenarios and the trajectories taken by our Transformer policy when turning left and right. Yellow star denotes position of the waypoint.}
  \label{fig:rules}
\end{figure}

\begin{figure}[tb]
  \centering
  \begin{subfigure}{0.4\linewidth}
    \includegraphics[width=\textwidth]{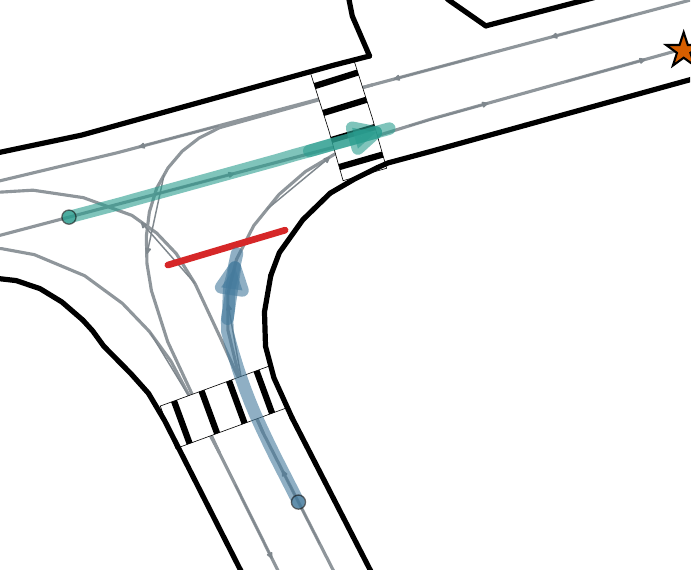}
    \caption{Blue vehicle slows down to yield to the green vehicle on the main road}
  \end{subfigure}
  \begin{subfigure}{0.4\linewidth}
    \includegraphics[width=\textwidth]{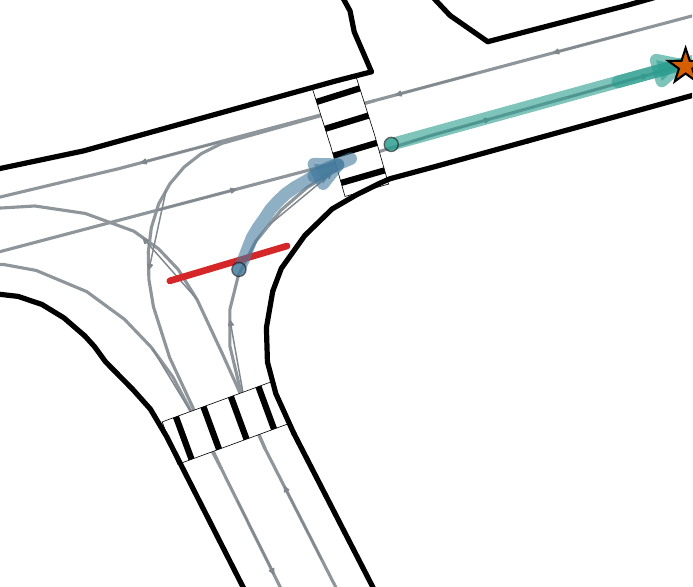}
    \caption{After green vehicle has passed, blue vehicle continues driving}
  \end{subfigure}
  \caption{Example of a synthetic scenario containing a stop line and the trajectories taken by agents. Yellow star shows the position of the waypoint for both agents.}
  \label{fig:rules-stopline}
\end{figure}

\begin{figure}[tb]
  \centering
  \begin{subfigure}{0.45\linewidth}
    \includegraphics[width=\textwidth]{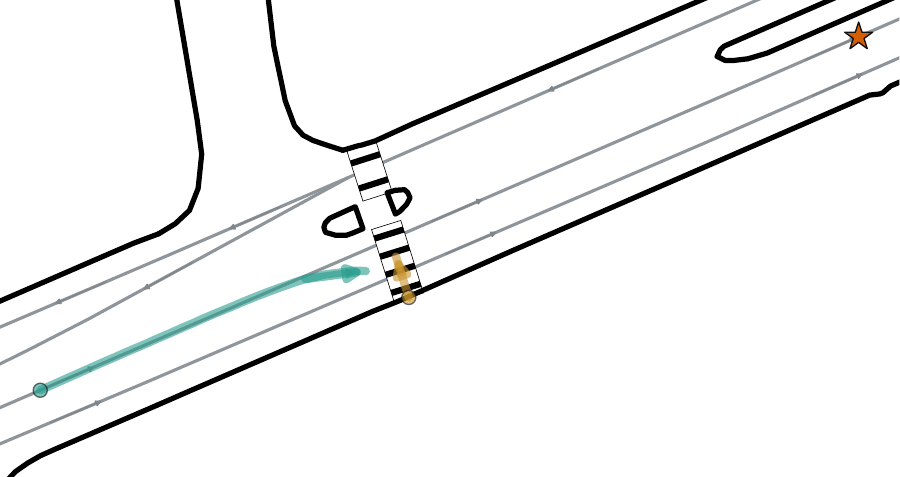}
    \caption{Agent navigates to the right lane to safely pass the pedestrian}
  \end{subfigure}
  \begin{subfigure}{0.45\linewidth}
    \includegraphics[width=\textwidth]{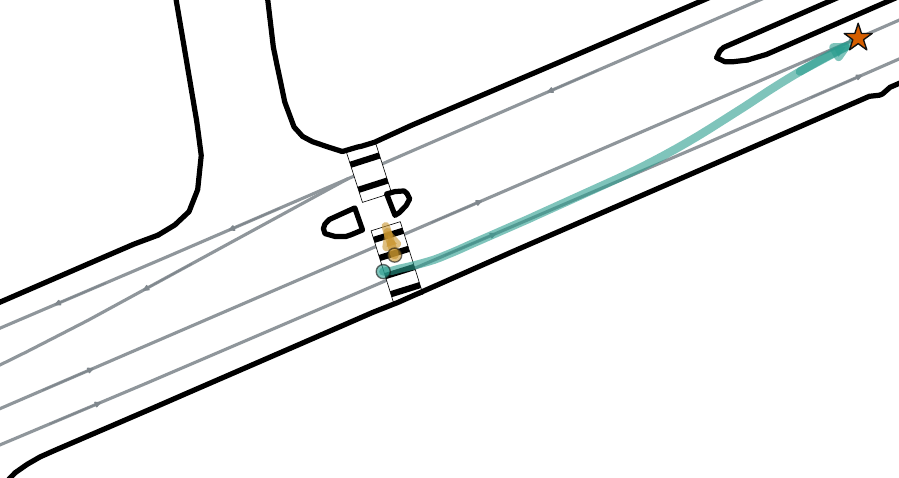}
    \caption{Agent navigates back to left lane after the crosswalk}
  \end{subfigure}
  \caption{Example of a synthetic scenario where pedestrian (in yellow) follows fixed trajectory along a crosswalk. Yellow star shows position of the waypoint for the agent.}
  \label{fig:rules-crosswalk}
\end{figure}

Next, we study whether the learned policy follows the traffic rules of the real world. To do this, we construct scenarios on the map of Tartu, let the policy drive them, and manually assess its behavior. Because of the manual effort involved, we restrict this evaluation to the Transformer model.

\subsubsection{Turning from the Correct Lane.} In real traffic, there are often dedicated lanes for making left and right turns. We construct 128 scenarios where a vehicle has to make a turn but starts in the wrong lane and must change lanes before turning. Among these 128 scenarios, we observe a single failure in which the vehicle starts the turn from the wrong lane. However, for left turns the policy changes lanes only 1--3 seconds before the turn itself. Interestingly, this is not the case for right turns: in around 90\% of cases, the policy changes lanes almost immediately and more gradually. \Cref{fig:rules-leftturn,fig:rules-rightturn} show example scenarios and the trajectories taken by the policy.

\subsubsection{Stop Lines.} As discussed in the CARLA evaluation, our policy lacks an explicit reward for yielding at stop lines. To measure their impact, we craft 63 scenarios where two vehicles approach an intersection or roundabout, one on the main road and one on a side road with a stop line in front of it. We set the initial positions so that the vehicles would collide if both continued at constant speed. In the evaluation runs, there are no collisions. In 59\% of cases, the vehicle on the main road yields to the vehicle on the side road, and in 41\% the vehicle on the side road yields to the vehicle on the main road. \Cref{fig:rules-stopline} shows an example scenario and the trajectories selected by the vehicles. We then repeat the evaluation with the stop lines hidden from the agent. The yielding behavior changes in only one scenario, in which the vehicle behind the stop line previously did not yield to the vehicle on the main road but now does. In all other scenarios, the yielding behavior remains exactly the same. These tests therefore provide no evidence that the policy uses the stop-line feature.

\subsubsection{Crosswalks.} First, we construct 64 scenarios where a pedestrian is waiting on the side of the road to cross at a crosswalk. In these scenarios, our policy stops in 39\% of cases to let the pedestrian cross the road. Then, we take the same scenarios and make the pedestrian walk over the crosswalk along a fixed trajectory. In those scenarios, our policy stops in front of the crosswalk in 14\% of cases to let the pedestrian pass. In 15\% of cases, the policy does not yield to the pedestrian and drives over the crosswalk in front of them. In the remaining 71\% of cases, the policy yields to the pedestrian by driving over the crosswalk behind them. \Cref{fig:rules-crosswalk} shows an example of such a case. Despite receiving no pedestrian-interaction reward, the policy yields in 39\% of waiting-pedestrian scenarios and 85\% of moving-pedestrian scenarios, but its behavior remains inconsistent.

\subsection{Behavior Diversity}

In \cref{sec:rewards}, we introduced randomized reward coefficients to encourage diverse agent behaviors, with the end goal of making the policy more robust. However, introducing such rewards does not by itself guarantee that the policy actually learns diverse behaviors. To verify this, we recorded certain events during training and analyzed how their frequencies vary with the reward conditioning coefficients. All results below are based on the last 10\% of the Transformer training run.

\begin{figure}[tb]
  \centering
  \begin{subfigure}{0.32\linewidth}
    \includegraphics[width=\textwidth]{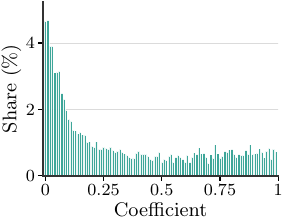}
    \caption{Off-road events vs off-road reward coefficient}
    \label{fig:diversity-offroad}
  \end{subfigure}
  \hfill
  \begin{subfigure}{0.32\linewidth}
    \includegraphics[width=\textwidth]{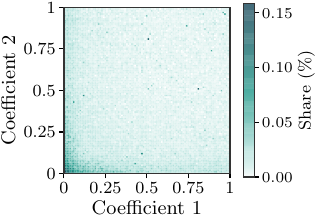}
    \caption{Collision events vs collision reward coefficient}
    \label{fig:diversity-collision}
  \end{subfigure}
  \hfill
  \begin{subfigure}{0.32\linewidth}
    \includegraphics[width=\textwidth]{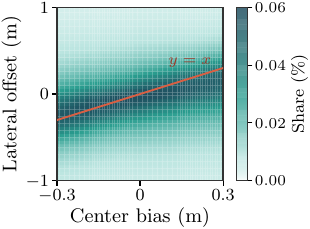}
    \caption{Lateral offset vs lane center bias}
    \label{fig:diversity-lateral}
  \end{subfigure}
  \caption{Diversity of agent behavior due to the conditioning coefficients. The data comes from the last 10\% of the Transformer training run.}
  \label{fig:diversity}
\end{figure}

\Cref{fig:diversity-offroad} visualizes the distribution of events where a vehicle goes off-road (collides with the road border) as a function of the off-road reward coefficient. Vehicles with a low coefficient clearly go off-road more frequently, but among vehicles with a coefficient larger than 0.25, the distribution becomes uniform and vehicles avoid the road border to the same extent. We hypothesize this is because beyond very low coefficient values, there is little benefit to colliding with the road border.

\Cref{fig:diversity-collision} shows the number of collisions as a function of the coefficients of the two vehicles involved in the crash. The heatmap shows two trends. First, in the region where both vehicles have a low collision coefficient, collisions become much more likely. For example, 8\% of collisions are between vehicles whose coefficients are both less than 0.1, despite such vehicles making up only 1\% of vehicle pairs on the road.

Second, two vehicles are more likely to collide as long as just one of them has a low collision coefficient, and the value of the other coefficient makes little difference. More concretely, in 49\% of collisions, one of the involved vehicles has a collision coefficient of less than 0.1. This could be because avoiding collisions is a collaborative effort, and there is only so much one driver can do.

Finally, \cref{fig:diversity-lateral} shows the observed lateral offset values (the vehicle's distance from the lane center) as a function of the lane center bias. The lane center bias represents the offset from the actual lane center that is treated as the lane center for that particular vehicle --- when the vehicle drives at this offset, it receives the highest lane center reward. As one would expect, the observed lateral offsets are closely clustered around the bias. For very high biases near 0.3 m, the lateral offsets deviate from the trend, likely because positive biases are to the right of the lane center and there may not be enough room to drive that far right due to road edges.

\section{Limitations and Future Work}

The ultimate goal of this line of research is to deploy the learned policy in real traffic in Tartu, Estonia. Our results highlight several obstacles on this path. The collision and off-road rates remain too high for real-world use, and the policies obey the traffic rules inconsistently by not yielding at stop lines or crosswalks and by reward hacking at traffic lights. We see two complementary approaches to closing this gap. The first is scale: training longer and training larger models. Both of our models have around 600K parameters, roughly a tenth of the size of the Gigaflow model and tiny by the standards of modern machine learning.

The second approach is to align the policy with the traffic code explicitly rather than relying on alignment to emerge with scale. As our results show, some behaviors desired in real traffic did not emerge naturally and may need more specific guidance to be learned, either via additional rewards or imitation learning. More sophisticated reward terms could be introduced into the self-play, such as a penalty for passing pedestrians who are waiting to cross the road. The traffic light reward in particular requires more careful modeling, \eg, an additional penalty for driving through an intersection while the light is red, so that crossing through the oncoming lane no longer avoids the penalty.

On the other hand, for some behaviors, specifying the reward may be non-trivial. For example, we have yet to come up with a general algorithm for determining whether a vehicle yielded at a stop line or not. To learn such behaviors, self-play could be combined with learning from human data, as demonstrated by Cornelisse~\etal~\cite{spicedselfplay}.

A more fundamental limitation of our current work is that it assumes perfect observation of the world. Since our deployment is limited to Tartu, assuming a perfect HD map is realistic. Perfect detection of other vehicles and objects, however, is not: some objects are hidden behind physical barriers such as walls, and detections inevitably contain errors. To work in practice, our models must therefore be trained under partial observability, with occlusions and detection noise included in the simulation.

\section{Conclusion}

Building on ideas from Gigaflow and using the PufferDrive training simulator, we trained Deep Sets and Transformer policies for autonomous driving through self-play without human driving data, with the end goal of deploying them in Tartu, Estonia. Although the policies were trained solely on a map of Tartu and without human driving data, we evaluated them on the CARLA and Waymax benchmarks, where they fell short of the performance of Gigaflow. Our analysis identified likely contributors this gap to several failure modes, including reward hacking at traffic light stop lines, disregarding stop signs, and relying too heavily on the cooperation of other vehicles. At the same time, the policies almost always turned from the correct lane in our constructed scenarios, negotiated intersections without collisions, and exhibited the diverse driving behaviors targeted by the reward conditioning. We see these results as a step towards deploying a self-play policy in real traffic, with scale, reward design, and partial observability as the main challenges ahead.

\bibliographystyle{splncs04}
\bibliography{main}
\end{document}